\documentclass[conference]{IEEEtran}
\usepackage{cite}
\IEEEoverridecommandlockouts
\usepackage{amsmath,amssymb,amsfonts}
\usepackage{algorithmic}
\usepackage{graphicx}
\usepackage{svg}
\usepackage{textcomp}
\usepackage{xcolor}
\usepackage{hyperref}
\usepackage{subcaption}
\usepackage{capt-of}  %
\usepackage{cuted}    %
\def\BibTeX{{\rm B\kern-.05em{\sc i\kern-.025em b}\kern-.08em
    T\kern-.1667em\lower.7ex\hbox{E}\kern-.125emX}}
\begin{document}

\title{CUDA-Accelerated Soft Robot Neural Evolution with Large Language Model Supervision\\
}
\author{Lechen Zhang,~\IEEEmembership{Student Member,~IEEE}
\thanks{L. Zhang was with the Department of Mechanical Engineering, 
Columbia University, New York, NY, 10027, USA email: lechen.zhang@columbia.edu}}
\maketitle

\begin{abstract}
This paper addresses the challenge of co-designing morphology and control in soft robots via a novel neural network evolution approach. We propose an innovative method to implicitly dual-encode soft robots, thus facilitating the simultaneous design of morphology and control. Additionally, we introduce the large language model to serve as the control center during the evolutionary process. This advancement considerably optimizes the evolution speed compared to traditional soft-bodied robot co-design methods. Further complementing our work is the implementation of Gaussian positional encoding - an approach that augments the neural network's comprehension of robot morphology. Our paper offers a new perspective on soft robot design, illustrating substantial improvements in efficiency and comprehension during the design and evolutionary process.
\end{abstract}


\section{Introduction}
In soft-bodied robots, the design process becomes complex due to the intricate interactions between the robot's morphology and control methods. Traditional approaches have regarded the co-design as a two-step process – focusing on evolving the controller for a fixed body morphology or alternating between optimizing robot morphology and control.

In contrast, our method seeks to co-design soft robot systems using an evolutionary algorithm integrated with a single genome. The proposed method dual encodes the morphological and control designs within the multilayer perceptron (MLP), utilizing the neural evolution method for innovative MLP generation. To address the drawbacks of the evolutionary algorithm, we utilized Large Language Model (LLM) supervision, enhancing evolutionary co-design speed while preserving diversity in the design generation. Our approach promises to improve soft robot design efficacy and efficiency, providing a foundation for advancements in the field.

\section{Method}
\begin{figure}[htbp]
\centering
\includeinkscape[width=0.50\textwidth]{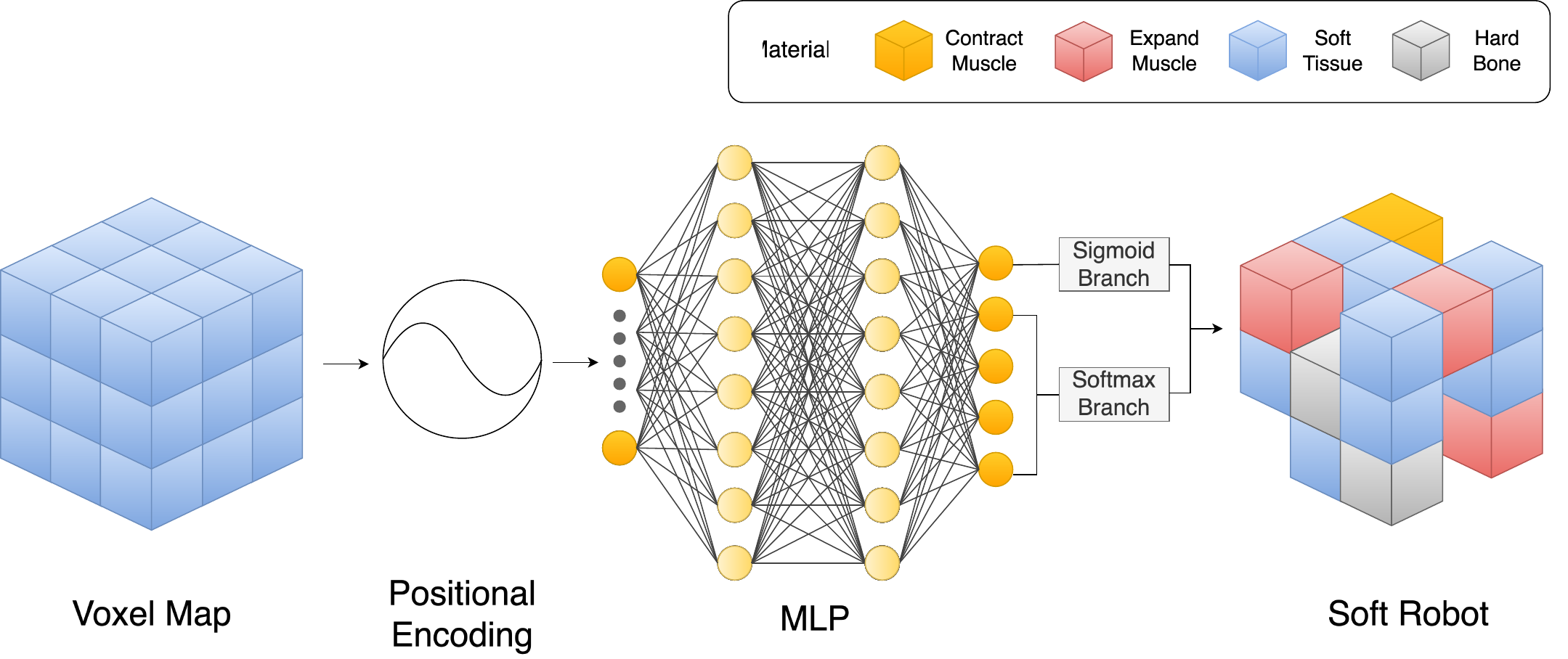}
\caption{A overview of the implicit dual encoding paradigm}
\label{fig1}
\end{figure}
\subsection{3D Voxel-Based Soft Robot}
Our method employs the 3D voxel-based soft robot model \cite{cheney2014unshackling} for robot generation driven by a spatial query model, which allows us to vary the presence and type of material at each voxel position. As illustrated in Fig. \ref{fig1}, choices for creating a robot's morphology include empty spaces, expanding and contracting muscles, and support structures made from soft tissue or hard bone material. This offers various design possibilities and flexibility, accommodating various application design requirements. The 3D voxel-based model simplifies the representation of complex structures and materials, facilitating the effective co-design of morphology and control.

\subsection{Gaussian Positional Encoding}
Inspired by recent work on neural rendering\cite{tancik2020fourier}, Gaussian positional encoding is utilized to map the spatial query input into higher dimensions. The encoding scheme can be mathematically represented as:

\begin{equation}
\gamma(v) = [\cos(2\pi Bv), \sin(2\pi Bv)]^T
\end{equation}

where each entry in \(B \in R^{m \times d}\) is drawn from the isotropic Gaussian distribution \(N(0, \sigma^2)\). The standard deviation, \(\sigma\), is selected by the hyperparameter sweep. This isotropic Gaussian distribution does not assume a strong prior on the signal's frequency spectrum, thereby aiding our model in better understanding the dynamics of the robot's morphology. This, in turn, significantly augments the efficiency of our co-design methodology.

\begin{figure}[htbp]
\centering
\includeinkscape[width=0.50\textwidth]{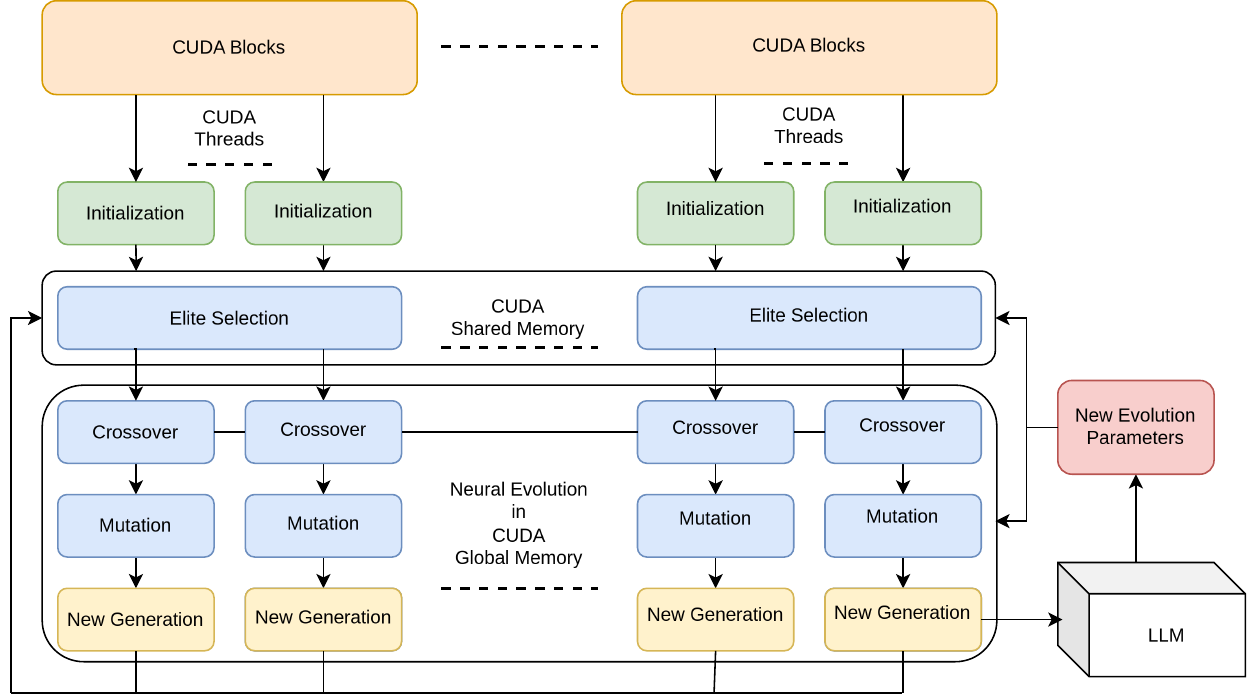}
\caption{A overview of CUDA-accelerated neural evolution framework}
\label{fig2}
\end{figure}

\subsection{CUDA-Accelerated Neural Evolution}
Our co-design methodology employs an evolutionary algorithm to tune the weights and biases within a population of MLPs and encourage the emergence of innovative MLP configurations. As illustrated in Fig. \ref{fig1}, the material score from the model's Softmax branch will determine the material category and the material weight from the model's Sigmoid branch will be multiplied by the material parameters shown in Table \ref{tab:hresult}, determine the result parameter of the input spatial position. To accelerate the simulation and neural evolution process, our simulator is built upon previous works that utilize Compute Unified Device Architecture(CUDA) to accelerate soft robot simulation\cite{austin2020titan}, \cite{xia2021legged}. Fig. \ref{fig2} illustrates how CUDA accelerates the evolution process. Each CUDA thread in the CUDA block will be responsible for the property update spring or mass inside one soft robot. And the crossover, mutation and elite selection process are also fully parallel in GPU. This approach enabled up to 7,892,537,853 spring-mass physics simulations per second within one NVIDIA RTX-3090 GPU.

\subsection{LLM Supervision}
Inspired by recent work that combines LLMs and Evolution Algorithms\cite{lehman2023evolution}, the typical limitations of evolutionary algorithms, including slow convergence and complex hyperparameter tuning, are addressed by using LLM supervision. Our research leveraged the capabilities of GPT-4-Turbo, which facilitates extensive text analysis, to guide the evolution of data and generate novel hyperparameters for neural evolution and soft robot generation. As illustrated in Fig. \ref{fig2}, the parameter and population 3 generations prior to the current generation are input to the LLM and generate the parameter for this generation. This approach enabled us to optimize the evolutionary process and explore innovative material parameters for enhanced soft robot design.

\section{Results}

\begin{figure}
\centering
\begin{subfigure}{.25\textwidth}
  \centering
  \includeinkscape[width=\textwidth]{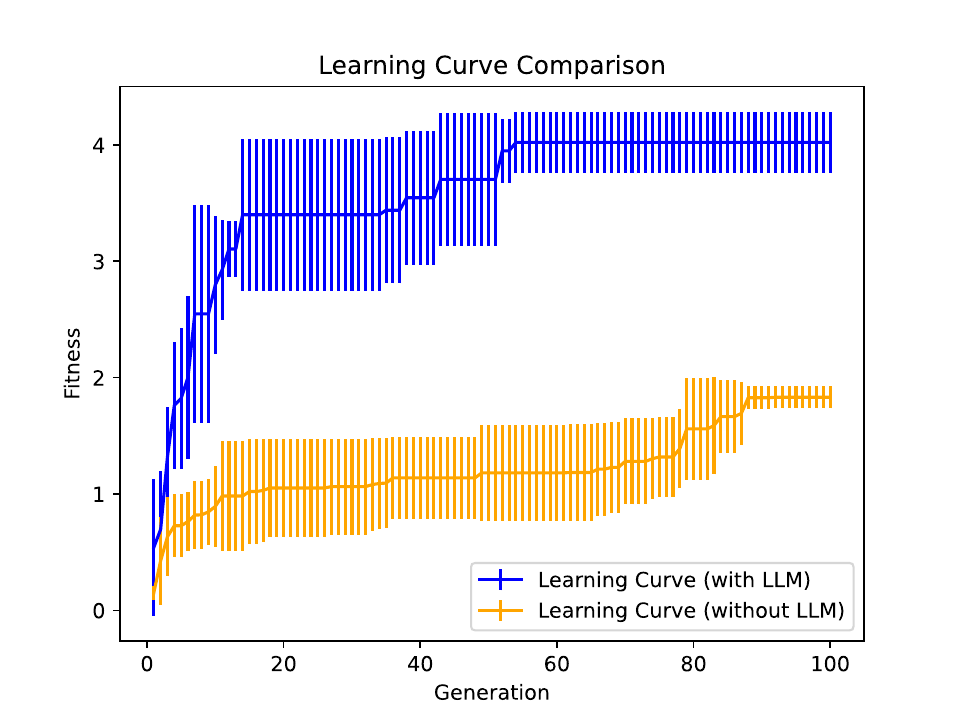}
  \caption{Learning Curve}
  \label{fig:sub3}
\end{subfigure}%
\begin{subfigure}{.25\textwidth}
  \centering
  \includeinkscape[width=\textwidth]{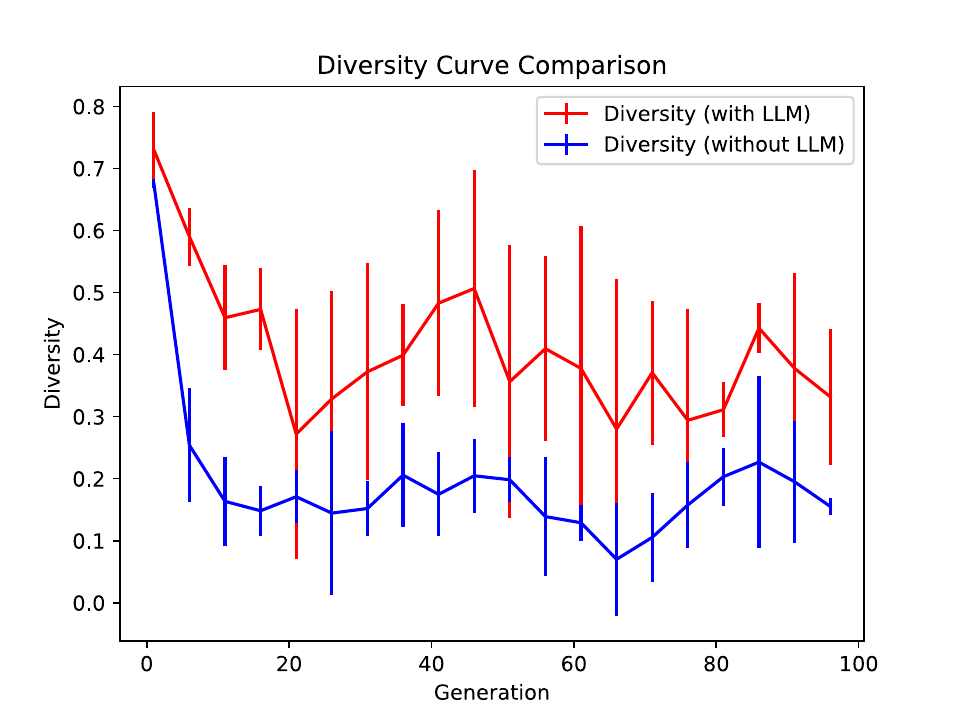}
  \caption{Diversity Curve}
  \label{fig:sub4}
\end{subfigure}
\caption{A comparison of neural evolution results with and without LLM supervision}
\label{fig:test1}
\end{figure}

\begin{figure}[htbp]
\centering
\includeinkscape[width=0.50\textwidth]{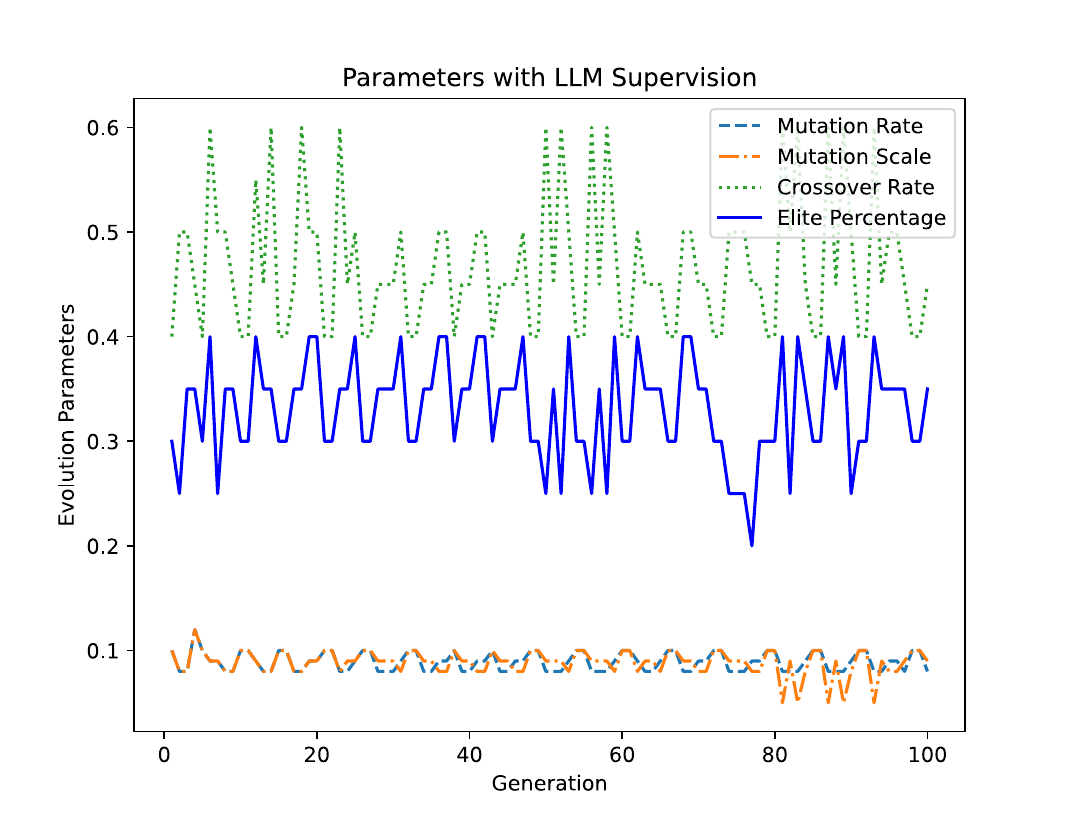}
\caption{How GPT-4-Turbo changes the hyperparameters}
\label{fig}
\end{figure}

\begin{figure}
\centering
\begin{subfigure}{.25\textwidth}
  \centering
  \includegraphics[width=\textwidth]{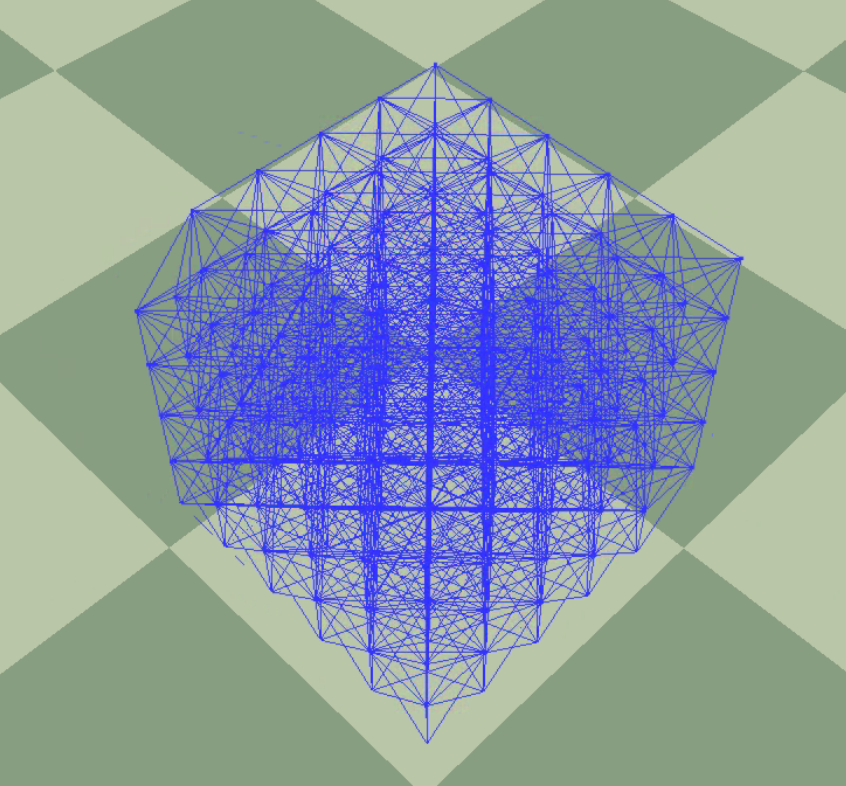}
  \caption{Soft robot at gen 0}
  \label{fig:sub1}
\end{subfigure}%
\begin{subfigure}{.25\textwidth}
  \centering
  \includegraphics[width=\textwidth]{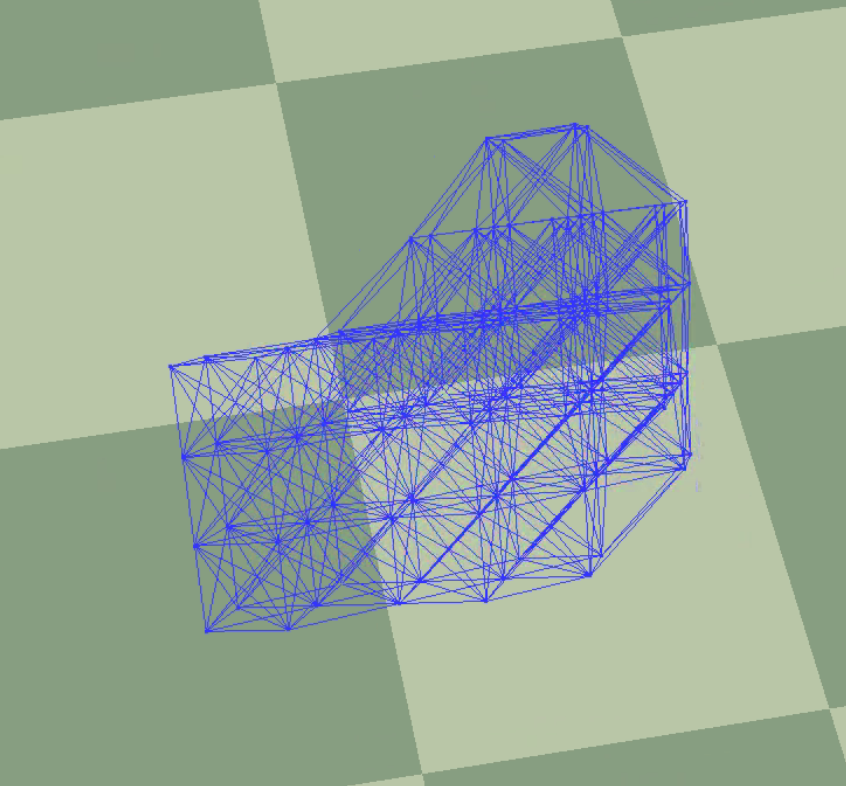}
  \caption{Soft robot at gen 100}
  \label{fig:sub2}
\end{subfigure}
\caption{A comparison of soft robot results at gen 0 and 100}
\label{fig:test2}
\end{figure}
To evaluate our model's effectiveness, we ran the evolutionary process under the same initial hyper-parameters shown in Table \ref{tab:param}, each conducted three times with and without the supervision of the LLM. Our results indicate a clear advantage of applying LLM supervision in our co-design methodology. As shown in Fig. \ref{fig:sub3}, using LLM helps maintain a high level of diversity within the evolved robot morphologies and control strategies. This diversity is critical to avoid premature convergence to sub-optimal solutions and to explore a broader space of possibilities. Additionally, as illustrated in Fig. \ref{fig:sub4} LLM supervision noticeably accelerates the rate at which fitness convergence is achieved, thus enabling more efficient and result-oriented co-design. The soft robot in initial population and final evolution results are illustrated in Fig. \ref{fig:test2}, rendered using the OpenGL library.

\begin{table}[ht] 
\caption{General Parameters} %
\centering %
\begin{tabular}{||c c||} 
 \hline
 Parameters & Value \\
 \hline\hline
  Generations & $100$ \\
 \hline
   Population Size & $30$ \\
 \hline
 Robot Size & $5*5*5$ \\
  \hline
  Repetitions & $3$\\
    \hline
  Initial Mutation Rate & $0.1$\\
    \hline
  Initial Mutation Scale & $0.1$\\
    \hline
  Initial Crossover Rate & $0.4$\\
    \hline
  Initial Elite Percentage & $0.3$\\
    \hline
\end{tabular}
\label{tab:param} 
\end{table} 

\begin{table}[ht] 
\caption{Simulation Parameters} %
\centering %
\begin{tabular}{||c c||} 
 \hline
 Parameters & Value\\
 \hline\hline
  \multicolumn{2}{|c|}{\textbf{Physics Parameters}}\\
  \hline
 Gravity & $9.81 m/s^{2}$\\
 \hline
 Mass per Indices & $0.1Kg$\\ 
 \hline
 Original Rest Length & $0.1m$\\
 \hline
  Simulation Time-Step & $10^{-5}s$\\
 \hline
  \multicolumn{2}{|c|}{\textbf{Material Parameters}}\\
 \hline
 Muscle Spring Constant & $2*10^{3}N/m$ \\
 \hline
 Soft Tissue Spring Constant & $10^{3}N/m$ \\
 \hline
 Hard Bone Spring Constant & $10^{4}N/m$ \\
 \hline
 Spring Damping Ratio & $0.1$\\
 \hline
 Muscle Max Amplitude & $0.25$ \\
 \hline
 Muscle Max Phase & $\pi$ \\
 \hline
 \multicolumn{2}{|c|}{\textbf{Plane Parameters}}\\
 \hline
 Elastic Coefficient & $10^{5} N/m$ \\
 \hline
 Damping Ratio & $0.1$ \\
 \hline
  Static Coefficient of Friction & $0.6$ \\
 \hline
   Kinetic Coefficient of Friction & $1.0$ \\
 \hline
\end{tabular}
\label{tab:hresult} 
\end{table}

\end{document}